\documentclass[conference]{IEEEtran}
\usepackage{spconf,amsmath,graphicx}
\usepackage{mathtools}

\usepackage{gensymb}

\newcommand\given[1][]{\:#1\vert\:}

\begin{document}

\title{3D HIGH-RESOLUTION CARDIAC SEGMENTATION RECONSTRUCTION FROM 2D VIEWS USING CONDITIONAL VARIATIONAL AUTOENCODERS}

\maketitle
\begin{abstract}
Accurate segmentation of heart structures imaged by cardiac MR is key for the quantitative analysis of pathology. High-resolution 3D MR sequences enable whole-heart structural imaging but are time-consuming, expensive to acquire and they often require long breath holds that are not suitable for patients. Consequently, multiplanar breath-hold 2D cines sequences are standard practice but are disadvantaged by lack of whole-heart coverage and low through-plane resolution. To address this, we propose a conditional variational autoencoder architecture able to learn a generative model of 3D high-resolution left ventricular (LV) segmentations which is conditioned on three 2D LV segmentations of one short-axis and two long-axis images. By only employing these three 2D segmentations, our model can efficiently reconstruct the 3D high-resolution LV segmentation of a subject. When evaluated on 400 unseen healthy volunteers, our model yielded an average Dice score of $87.92 \pm 0.15$ and outperformed competing architectures (TL-net, Dice score = $82.60 \pm 0.23$, $p=2.2 \cdot 10^{-16}$). 
\end{abstract}
\begin{keywords}
Cardiac MR, Variational Autoencoder, 3D Segmentation Reconstruction, Deep Learning.
\end{keywords}
\section{Introduction}
\label{sec:intro}

Cardiac magnetic resonance (CMR) is the gold-standard technique for assessment of cardiac morphology. Conventional practice is to acquire a stack of breath-hold 2D image sequence in the left ventricular (LV) short axis supplemented by long axis image sequence in prescribed planes to enable reproducible volumetric analysis and diagnostic assessment \cite{alfakih2004assessment}. Disadvantages of this approach for whole-heart segmentation are low through-plane resolution, misalignment between breath-holds and lack of whole-heart coverage. High-resolution 3D image sequences address some of these issues, but also have disadvantages in terms of long acquisition times, relatively low in-plane resolution and lack of clinical availability. However, high-resolution 3D segmentations proved to be crucial for the construction of integrative statistical models of cardiac anatomy and physiology and disease characterization \cite{biffibio, bai2015bi}. For these reasons, a method to reconstruct a 3D high-resolution segmentation from routinely-acquired 2D cines could be highly beneficial - offering high resolution phenotyping robust to artefact in large clinical populations with conventional imaging.

The reconstruction of 3D anatomical structures from a limited number of 2D views has been previously studied via deformable statistical shape models \cite{whitmarsh2011reconstructing}. However, these methods require complex reconstruction procedures and are very computationally-intensive. In recent years, with the advent of learning-based approaches, and in particular of deep learning, a number of alternative strategies have been proposed. The TL-embedding network (TL-net) consists of a 3D convolutional autoencoder (AE) which learns a vector representation of the 3D geometries, whereas a second convolutional neural network attached to the latent space of the AE maps 2D views of the same object to the same vector representation \cite{girdhar2016learning}. More recently, \cite{cerrolaza20183d} proposed a convolutional conditional variational autoencoder (CVAE) architecture for the 3D reconstruction of the fetal skull from 2D ultrasound standard planes of the head.  Finally, \cite{biffi2018learning} showed how a convolutional variational autoencoder (VAE) can learn a shape segmentation model of left ventricular (LV) segmentations and how the learned latent space can be exploited to accurately identify healthy and pathological cases and generate realistic segmentations unseen during training.

\begin{figure*}
\begin{center}
  \includegraphics[width=0.9\textwidth]{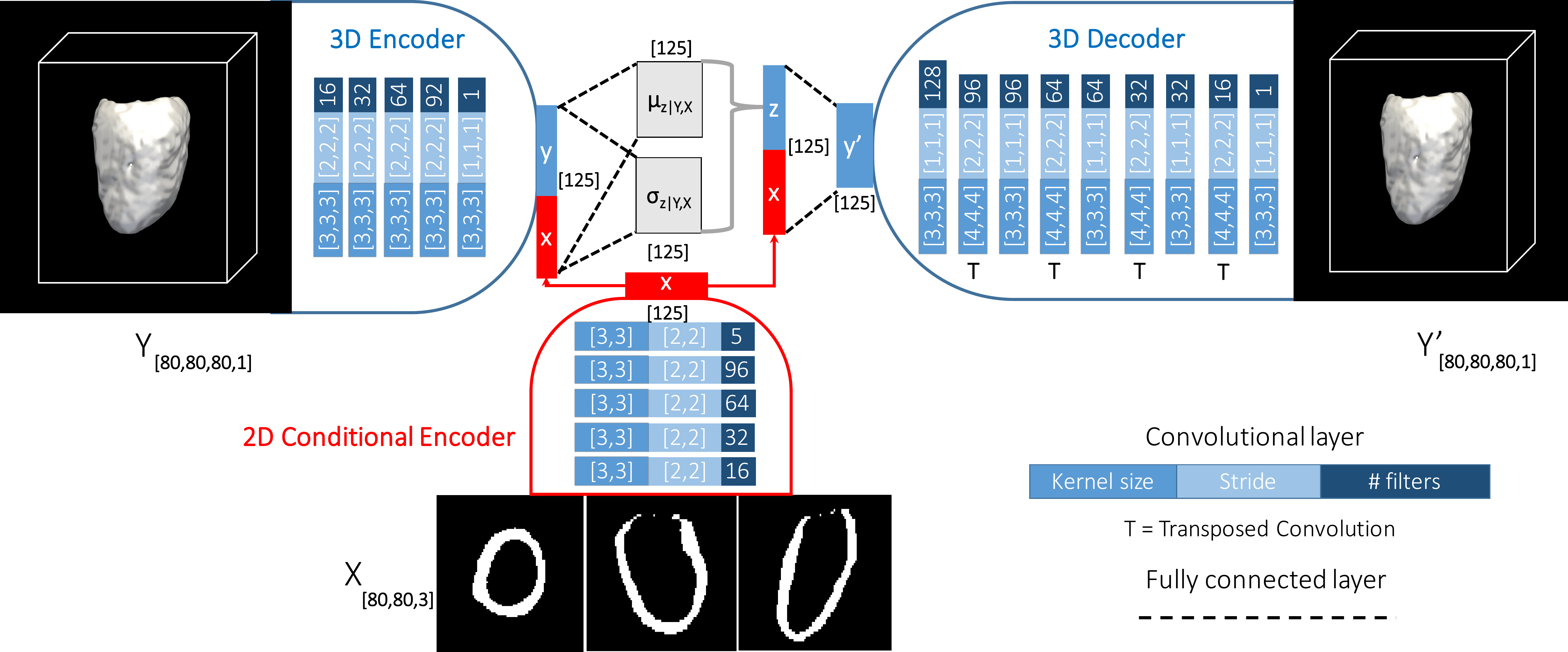}
  \caption{The proposed conditional variational autoencoder (CVAE) architecture.}
  \label{fig:arch}
\end{center}
\end{figure*}

In this work, we present a CVAE architecture that reconstructs a high-resolution 3D segmentation of the LV myocardium from three segmentations of 2D standard cardiac views (one short-axis and two long-axis). Moreover we show how the proposed model naturally produces confidence maps associated to each reconstruction, unlike deterministic models, thanks to its generative properties.

\section{MATERIALS AND METHODS}
\label{sec:method}
\subsection{3D Cardiac Image Acquisition and Segmentation}
A high-spatial resolution 3D balanced steady-state free precession cine MR image sequence was acquired from 1,912 healthy volunteers of the UK Digital Heart Project at Imperial College London using a 1.5-T Philips Achieva system (Best, the Netherlands) \cite{bai2015bi}. Left and right ventricles were imaged in their entirety in a single breath-hold (60 sections, repetition time 3.0 ms, echo time 1.5 ms, flip angle 50$\degree$, field of view $320 \times 320 \times 112 $ mm, matrix $160 \times 95$, reconstructed voxel size $1.2 \times 1.2 \times 2$ mm, 20 cardiac phases, temporal resolution 100 ms, typical breath-hold 20 s). For each subject, a 3D high-resolution segmentation of the LV was automatically obtained using a previously reported technique employing a set of manually annotated atlases \cite{bai2015bi}. In this work, only the end-diastolic (ED) frame was considered.  

\subsection{Conditional Variational Autoencoder Architecture}
The outline of the CVAE architecture we propose is shown in Fig. 1. We aim at reconstructing a 3D high-resolution LV segmentation $\bf{Y}$ from $i$ segmentations obtained in as many 2D views $\bf{X}$ $= \{ X_i \; | \; i = 1, 2, 3 \}$. We aim to learn from the training data a conditional generative model $P(\bf{Y}|\bf{X})$ by means of a $d$-dimensional latent distribution $\bf{z}$ and a low-dimensional representation $\bf{x}$ of the views $\bf{X}$. In this work we use a single 2D convolutional neural network (CNN) to encode the 2D views $\bf{X}$ in a low-dimensional representation $\bf{x}$. An alternative encoding strategy was proposed in \cite{cerrolaza20183d}, using a separate branch for each conditional input of the model. However, whilst this latter approach proved efficient when the views suffer from large inconsistencies or variability (e.g., free-hand ultrasound scans), we can notably reduce the model complexity by combining the views $\bf{X}$ as a unique three-channel input as these are consistently acquired in clinical routine.

Directly inferring $P(\bf{Y}|\bf{X})$ is impractical as it would require sampling a large number of $\bf{z}$ values. However, variational inference allows us to approximate $P(\bf{Y}|\bf{X})$ by introducing a high-capacity function $\bf{Q}(\bf{z}|\bf{Y},\bf{X})$ which gives us a distribution over $\bf{z}$ values that are likely to produce $\bf{Y}$. Hence we can learn $P(\bf{Y}|\bf{X})$ by minimizing the following objective:
$$ log(P(\bf{Y}|\bf{X})) - D_{KL}[Q(\bf{z}|\bf{Y},\bf{X}) || P(\bf{z}\given\bf{Y},\bf{X})] = $$
\begin{equation} 
E_{\bf{z}  \sim \bf{Q}}[log P(\bf{Y} |\bf{z,X})] - D_{KL}[Q(\bf{z}|\bf{Y},\bf{X}) || P(\bf{z}\given\bf{X})]
\end{equation}
where $D_{KL}$ represents the Kullback-Leibler (KL) divergence of two distributions (full mathematical derivation of the equation can be found in \cite{doersch2016tutorial}). The encoding function $\bf{Q}(\bf{z}|\bf{Y},\bf{X})$ can be modelled as a Gaussian distribution parametrized by $\bf{\mu_{z|Y,X}}$ and $\bf{\sigma_{z|Y,X}}$ vectors. These two vectors can be learned by encoding the input 3D segmentation $\bf{Y}$ we want to reconstruct via a 3D CNN to a set of features $\bf{y}$, which are then concatenated together with the lower dimensional representation $\bf{x}$ of the views $\bf{X}$. By concatenating $[\bf{x},$ $\bf{y}]$ with a fully connected neural network to $\bf{\mu_{z|Y,X}}$ and $\bf{\sigma_{z|Y,X}}$ we can thus learn $\bf{Q}(\bf{z}|\bf{Y},\bf{X})$. 

If $\bf{Q}(\bf{z}|\bf{Y},\bf{X})$ is modelled by a sufficiently expressive function, then this function will match the real  $P(\bf{z} \given \bf{Y},\bf{X})$ and the $D_{KL}[Q(\bf{z}|\bf{Y},\bf{X})  || P(\bf{z}\given\bf{Y},\bf{X})]$ term in (1) will be zero. Therefore optimizing the right side of (1) will correspond to optimizing $P(\bf{Y}|\bf{X})$. In this work, the first term of the right side of (1) is computed as the Dice score (DSC) between $\bf{Y}$ and its reconstruction $\hat{\bf{Y}}$, which is the output of the generative model. The second term in (1) can be computed in a closed form if we assume its prior distribution to be $\mathcal{N}(\bf{0}, \bf{1})$, a $d$-dimensional normal distribution with zero mean and unit-standard deviation, and where $d$ is the number of dimensions of the latent space. Therefore the loss function we optimize becomes $\mathcal{L} = DSC(\bf{Y}, \hat{\bf{Y}}) + \alpha \; D_{KL}[\bf{z}||\mathcal{N}(0,1)]$. 

\subsection{Experimental Setup and Network Training}
In this work, we mimicked the two long-axis and the one short-axis views acquired in a routine acquisition with the following steps: (1) we rigidly aligned all the ground truth 3D high-resolution segmentations by performing landmark-based and subsequent intensity-based rigid registration; (2) we kept only the LV myocardium label and we cropped and padded the segmentations to [x = 80, y = 80, z = 80, t = 1] dimension using a bounding box centered at the centre of mass of the LV myocardium; (3) we sampled three orthogonal views passing through the centre of each segmentation (an example is shown in Fig. 1). Thanks to this process we extracted three 2D views showing the same three LV sections consistently for all subjects. In the following experiments, the ground truth 3D high-resolution segmentations and their corresponding 2D views were kept all in the same reference space. Inter-subject pose variability will be addressed in future work, potentially with a simple data augmentation strategy.

The dimension $d$ of the latent space was fixed to 125 as values smaller than 100 provided less accurate results, while above 125 no further improvements were observed. The dimensionality of the low dimensional representation $\bf{x}$ was kept equal to the dimensionality of $\bf{z}$ to guarantee a balanced contribution to the generative model. Simulations for different values of the parameter $\alpha$ in the loss function were performed: low values of $\alpha$ ($\alpha<0.5$) provided better reconstruction results on the training data at the expenses of a strong deviation from normality of the latent space distribution (KL term not converging) causing overfitting. Higher values of $\alpha$ ($\alpha>2$) penalized the reconstruction term in favour of a strictly normal latent space, hence providing poorer reconstruction accuracy. In this work we set $\alpha=1$ as this provided good reconstruction accuracy and convergence of the KL term. 

Experiments were performed with different numbers of views $X_i$ as conditions for the proposed model. In particular, referring to the first long-axis view as 1, the second long-axis view as 2 and the short-axis view as 3, we performed the training using either only one view (which we will indicate as CVAE\_1), or a combination or two views (CVAE\_12, CVAE\_23, CVAE\_13), or all the three views (CVAE\_123). We have also studied the feasibility of training a 2D AE to reconstruct the 3 views and used its encoder as a pre-trained conditional encoder (pCVAE\_123). Moreover, the reconstruction capability of the proposed architecture was compared with the one of the TL-net \cite{girdhar2016learning}. Finally, we compared the reconstruction obtained by a VAE with $\bf{z}$=$0$ (VAE\_0) to all our test segmentations, as this represents the best segmentation that the generative model can reconstruct when no information is provided to it. Results obtained with an autoencoder (AE) are also reported since this model yielded better results than different VAEs with distinct $\alpha$ values as it only optimizes the reconstruction accuracy. All the models share the same 3D encoder and decoder architectures.

The dataset was split into training, evaluation and testing sets consisting of 1362, 150 and 400 subjects respectively. Data augmentation included rotation around the three orthogonal axis with rotation angles randomly extracted from a normal distribution $\mathcal{N}(0,6\degree)$ and random closing and opening morphological operations. All the networks were implemented in Tensorflow and training was stopped after 300k iterations, when the total validation loss function had stopped improving (approximately 42 hours per network on an NVIDIA Tesla K80 GPU), using stochastic gradient descent with momentum (Adam optimizer, learning rate = $10^{-4}$) and batch size of 8. During testing, the 3D encoder branch was disabled and the reconstruction were obtained by setting the latent variables $\bf{z}$ $= \bf{0}$.

\begin{center}
\begin{table}[t!]
\resizebox{8.6cm}{!}{
\begin{tabular}{c|c|c|c}
Model     & DSC          & Hausd.  [mm]    & MassDiff {[}\%{]} \\
\hline
VAE\_0       & 65.48 $\pm$ 0.38 & 9.32 $\pm$ 0.06 & 35.37 $\pm$ 0.70 \\
\hline
CVAE\_1   & 78.08 $\pm$ 0.33 & 5.29 $\pm$ 0.04 & 3.94 $\pm$ 0.38\\
\hline
CVAE\_23  & 82.90 $\pm$ 0.21 & 4.43 $\pm$ 0.04 & 3.93 $\pm$ 0.19\\
CVAE\_12  & 85.21 $\pm$ 0.20 & 4.46 $\pm$ 0.04 & 3.73 $\pm$ 0.19\\
CVAE\_13  & 83.18 $\pm$ 0.18 & 4.77 $\pm$ 0.04 & 3.69 $\pm$ 0.19\\
\hline
\bf{CVAE\_123}    & \bf{87.92 $\pm$ 0.15} & \bf{3.99 $\pm$ 0.03} & \bf{2.70 $\pm$ 0.14}\\
pCVAE\_123& 87.63 $\pm$ 0.16 & 4.04 $\pm$ 0.04 & 3.05 $\pm$ 0.16\\
TL\_net   & 82.60 $\pm$ 0.23 & 4.66 $\pm$ 0.04 & 3.85 $\pm$ 0.19 \\
\hline
AE     & 90.45 $\pm$ 0.12 & 3.46 $\pm$ 0.03 & 1.50 $\pm$ 0.10  \\
\hline
\end{tabular}
}
\caption{Reconstruction metrics together with their standard error of the mean for all the studied models.}
\label{table}
\end{table}
\end{center}
\section{RESULTS AND DISCUSSION}
\label{sec:results}
\subsection{Accuracy of 3D Reconstruction}
Table 1 shows the reconstruction accuracy in terms of 3D Dice score, 2D slice-by-slice Hausdorff distance and LV mass difference between 3D high-resolution segmentations (ground truth and reconstructed ones) for all the studied architectures. LV mass is an important clinical biomarker, therefore we have estimated for each reconstruction its percentage difference in mass with the ground truth. The results indicate that the reconstruction accuracy decreases when views are removed. From the experiments with two views we can also infer how different views have different importance. In particular, the short-axis view seems to have the smallest impact on the reconstruction accuracy. This could be motivated by the fact that the long-axis views contain more information about the regional changes in curvature of the LV, which strongly influences the Dice Score. The results reported in Table 1 also show how our architecture significantly outperforms the TL-net by a large amount ($p=2.2 \cdot 10^{-16}$), and how the pre-training of the 2D CNN encoder network did not help to achieve better results. Finally, we can observe that the mass difference is systematically overestimated by a small amount that decreases with the number of views provided. We believe this a consequence of using the Dice score in the loss function. On the other hand, models trained using cross entropy in the loss function yielded a systematic underestimation of the mass, often with reconstructions with missing LV apex, as this loss term tends to favour the background instead of the myocardium.  

\begin{figure}[t!]
\begin{minipage}[b]{0.9\linewidth}
  \centerline{\includegraphics[width=8.5cm]{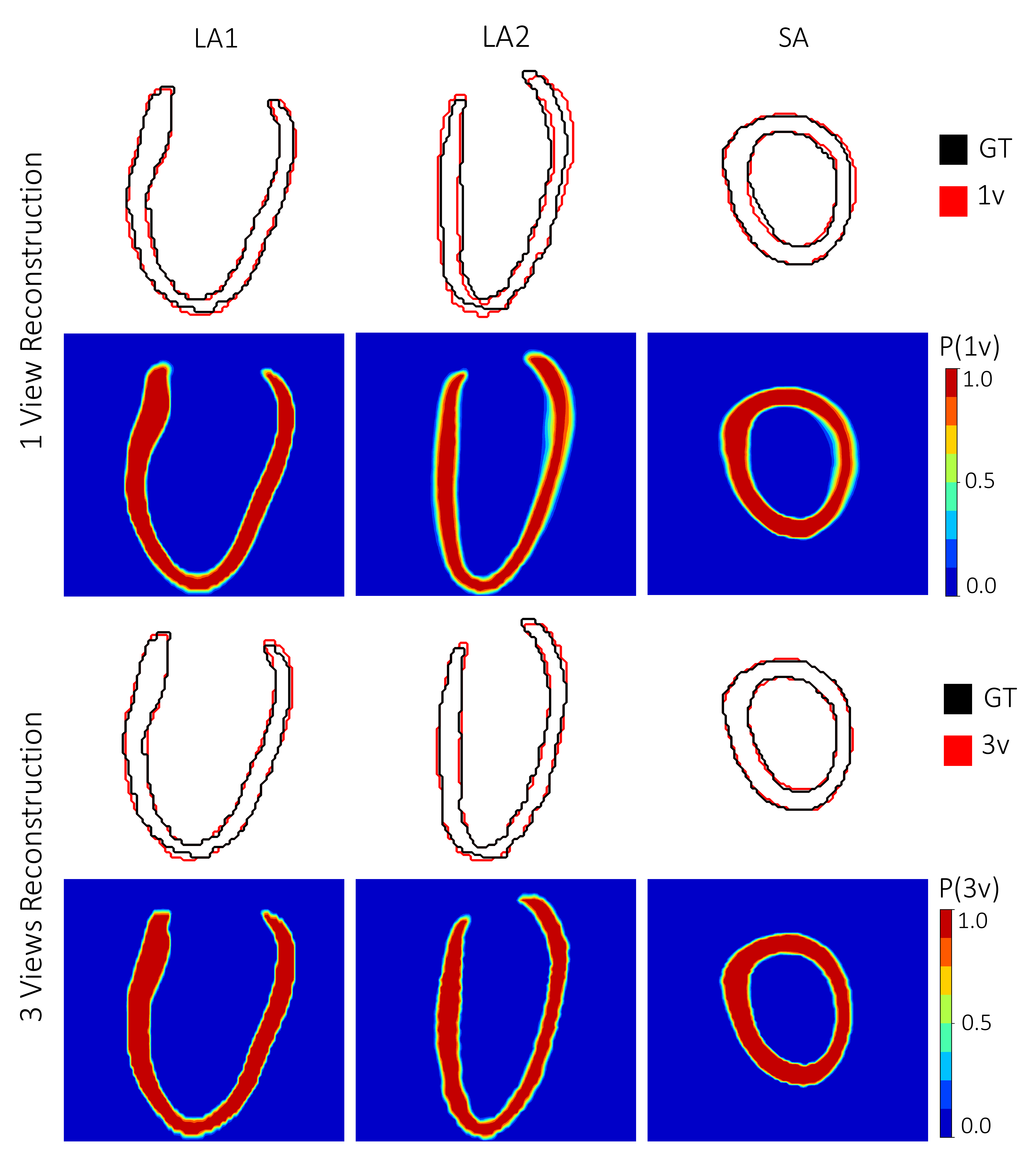}}
    \caption{First and third rows, reconstructed segmentation obtained with one and three views (in red, 1v and 3v) overlaid onto the ground truth segmentation (in black, GT) for one random subject. Second and fourth rows, confidence maps for the reconstruction with one and three views - $P(1v)$ and $P(3v)$. First and second columns, long-axis views (LA1 and LA2). Third column, short-axis (SA) view.}
  \label{fig:vi}
\end{minipage}
\end{figure}

\subsection{Visualisation and Uncertainty Estimation}
In the first and third rows of Fig.\ref{fig:vi} we report the reconstructed segmentations obtained with one and three views (in red) overlaid onto the ground truth segmentation (in black) for one subject of the testing dataset (with DSC 0.80 and 0.89, respectively). In the second and fourth rows we instead report the confidence maps obtained for the reconstruction with one and three views - $P(1v)$ and $P(3v)$. These maps have been obtained by sampling N times ($N=1,000$) $\bf{z}$ from $\bf{\mathcal{N}(0,1)}$ to reconstruct N segmentations from the same set of views $\bf{X}$. Unlike deterministic architectures (such as the TL-net), by averaging these maps we can compute the probability of each voxel to be labelled as LV myocardium, providing to clinicians a richer and more intuitive interpretation of the reconstruction. It can be seen in Fig. \ref{fig:vi} how the confidence map obtained with only 1 view has greater uncertainty than the one obtained with 3 views, which instead shows lower variability. Moreover, the amount of uncertainty in the $P(1v)$ map for the long-axis view 1 is less than for the other two views, as this view was the one provided to the network as condition. Interestingly, in the reconstruction with one view the areas with more uncertainty correspond to the areas where there is less overlap with the ground truth, i.e. the areas where the network is less accurate in predicting the shape.

\section{CONCLUSIONS}
\label{sec:conclusions}
In this paper we present the first deep conditional generative network for the reconstruction of 3D high-resolution LV segmentations from three segmentations of 2D orthogonal views. The reported results show the potential of this class of models to provide better quantitative cardiac models from sparse data. Future work will focus on using real standard long-axis views (instead of the simulated ones in this work), on reconstructing multiple structures and on extending the proposed framework to pathological datasets, for which acquiring breath-hold sequences is even more challenging.

\section*{\center \small{ACKNOWLEDGMENTS}}
The research was supported by the British Heart Foundation (NH/17/1/32725, RE/13/4/30184); National Institute for Health Research (NIHR) Biomedical Research Centre based at Imperial College Healthcare NHS Trust and Imperial College London; Academy of Medical Sciences Grant (SGL015/1006), and the Medical Research Council, UK. 

\newpage

\bibliographystyle{IEEEbib}

\bibliography{strings,refs}

\end{document}